\documentclass[lettersize,journal]{templete/IEEEtran}
\usepackage{amsmath,amsfonts}
\usepackage{algorithmic}
\usepackage{algorithm}
\usepackage{array}
\usepackage{CJKutf8}
\usepackage[caption=false,font=normalsize,labelfont=sf,textfont=sf]{subfig}
\usepackage{textcomp}
\usepackage{stfloats}
\usepackage{url}
\usepackage{verbatim}
\usepackage{graphicx}
\usepackage{cite}
\usepackage{amsmath,amssymb} 
\usepackage{mathrsfs}
\usepackage{pifont} 
\usepackage{bm}
\usepackage{algorithm}
\usepackage{algorithmic}
\usepackage{multirow}
\usepackage[colorlinks,
            linkcolor=red,
            urlcolor=blue,
            anchorcolor=red,
            citecolor=black]{hyperref}
\usepackage{colortbl}  
\usepackage{array}   
\definecolor{gray}{gray}{.9}
\usepackage{newfloat}
\usepackage{listings}
\DeclareCaptionStyle{ruled}{labelfont=normalfont,labelsep=colon,strut=off} 
\floatstyle{ruled}
\newfloat{listing}{tb}{lst}{}
\floatname{listing}{Listing}

\begin{document}
\begin{CJK}{UTF8}{gbsn}




\title{StructToken : Rethinking Semantic Segmentation with Structural Prior}
\author{ \normalsize
    Fangjian Lin\textsuperscript{\rm 1, \rm 2, \rm $\star$},
    Zhanhao Liang\textsuperscript{\rm 2,\rm 3, \rm $\star$},
    Sitong Wu\textsuperscript{\rm 4},
    Junjun He\textsuperscript{\rm 2},
    Kai Chen\textsuperscript{\rm 2, \rm 5},
    Shengwei Tian\textsuperscript{\rm 1, $\dag$}

\thanks{
\textsuperscript{$\star$}  Equal contributions.
\textsuperscript{$\dag$} Corresponding author.
\textsuperscript{\rm 1} School of Software, Xinjiang University, Urumqi, China.
\textsuperscript{\rm 2} Shanghai AI Laboratory, Shanghai, China.
\textsuperscript{\rm 3} Beijing University of Posts and Telecommunications.
\textsuperscript{\rm 4} Baidu Research.
\textsuperscript{\rm 5} SenseTime Research.
}}
     

\maketitle

\begin{abstract}
In previous deep-learning-based methods, semantic segmentation has been regarded as a static or dynamic per-pixel classification task, \textit{i.e.,} classify each pixel representation to a specific category.
However, these methods only focus on learning better pixel representations or classification kernels while ignoring the structural information of objects, which is critical to human decision-making mechanism.
In this paper, we present a new paradigm for semantic segmentation, named structure-aware extraction. Specifically, it generates the segmentation results via the interactions between a set of learned structure tokens and the image feature, which aims to progressively extract the structural information of each category from the feature.
Extensive experiments show that our StructToken outperforms the state-of-the-art on three widely-used benchmarks, including ADE20K, Cityscapes, and COCO-Stuff-10K.
\end{abstract}


\begin{figure*}[tp]
\centering
\includegraphics[width=1\linewidth]{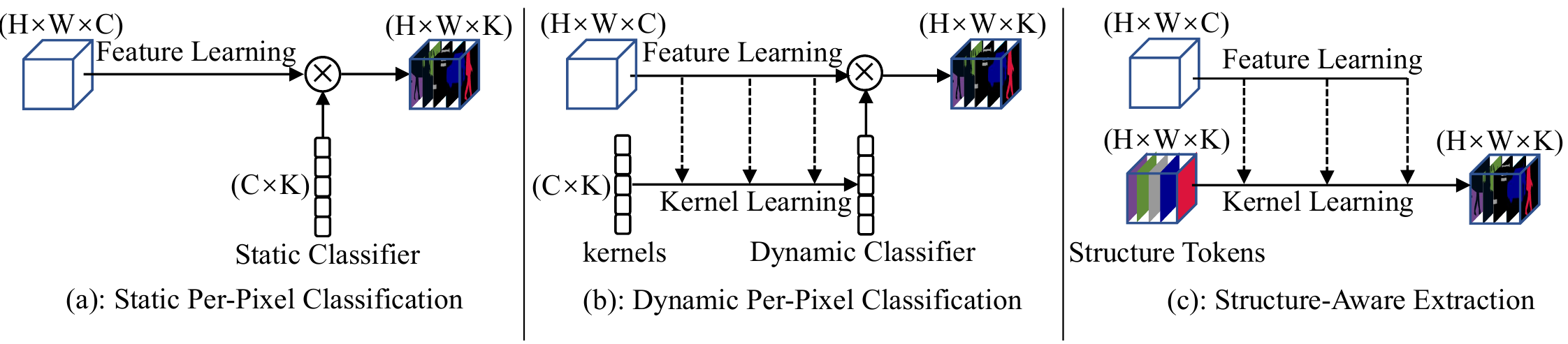} 
\caption{Comparison with three semantic segmentation paradigms. In (a), the segmentation results are obtained by the multiplication between the final feature map and a static classifier, where the classifier is fixed after training. 
By contrast, (b) further updates the initial kernels according to the image content to generate a dynamic classifier for each input image.
In our (c), it learns a set of structure tokens, and gradually extract information from the feature map to update structure tokens. The final structure tokens can be regarded as the segmentation results directly. $C$ and $K$ represent the number of channels and categories, respectively.}
\label{diff}
\end{figure*}

\section{Introduction}


\IEEEPARstart{W}ith the development of self-driving technology \cite{AutomaticDriving}, human-computer interaction \cite{HumanComputerInteraction}, and augmented reality \cite{AugmentedReality}, semantic segmentation has attracted more and more attention. 

Since the deep-learning era, semantic segmentation has been mainly formulated as a per-pixel classification task, that is, classifying each pixel to a specific category via a learned classifier (such as a 1$\times$1 convolution). 
According to the property of classifier, previous works can be categorized as two paradigms: \textit{static per-pixel classification} and \textit{dynamic per-pixel classification}.
As shown in Figure \ref{diff}\textcolor{red}{a}, for the static per-pixel classification paradigm, the classifier is fixed after the training process. The methods following this paradigm mainly focused on how to learn better representation for each pixel through context modeling \cite{PSPNet,v3+,SemanticFPN,Upernet,OCRNet} or automatic architecture design \cite{nas_seg1,nas_seg2,autodeeplab}.
As the static classifier learned from the dataset can be regarded as a comprehensive representation of each class, which may not be consistent with the representation of each object in every image, some recent works \cite{Segmenter,Maskformer,Mask2former} proposed to dynamically learn a classifier for different inputs according to their own contents. As shown in Figure \ref{diff}\textcolor{red}{b}, the initial kernel is updated by the image feature, resulting in a dynamic classifier more adaptive to the current input.

In these two paradigms, the entire decoder is dedicated to learning better features (including precise semantics and details) and a more robust classifier, and the segmentation decision is only performed in the final segmentation head via the per-pixel classification.
However, from the human perspective, the decision-making process of semantic segmentation presents a different pattern.
In particular, based on the underlying knowledge of category-wise structural information (such as texture, shape and spatial layout), human beings first determine the rough area of each category and then gradually refine it, rather than paying close attention to all the image information at first and final performing the classification at one time. 
This motivates us to explore whether a paradigm more in line with the human decision-making process is better than the previously popular per-pixel classification for semantic segmentation.



In this paper, we design a human-like paradigm for semantic segmentation, named structure-aware extraction. 
To simulate human knowledge, we define a set of learnable structure tokens, each of which is expected to model the implicit structural information of one category.
As shown in Figure \ref{diff}\textcolor{red}{c}, given the image feature, the structure tokens gradually extract information from the image feature. 
Qualitative visualization shows that the structural information becomes more and more explicit during progressive extraction. Thus, the refined structure tokens of the final layer can be directly regarded as the segmentation result.
Obviously, our paradigm is similar to the human-like process that uses structural knowledge to make rough discrimination first and then gradually perform refinement.

Following our structure-aware extraction paradigm, we further design a semantic segmentation network, named StructToken, to evaluate the effectiveness of our paradigm.
As mentioned above, the extraction aims to construct the mapping from channel slices of image features to those of structure tokens.
We instantiate the extraction operation in both content-agnostic and content-related manners, resulting in three different implementations, namely point-wise extraction (PWE), self-slice extraction (SSE) and cross-slice extraction (CSE). The corresponding three variants are denoted as StructToken-CSE, StructToken-SSE and StructToken-PWE, respectively.
Specifically, PWE and SSE apply point-wise convolution and channel-wise self-attention to the concatenation of the image feature and structure tokens, respectively. CSE performs channel-wise cross-attention between structure tokens and image feature, where the former is used as query and the latter is treated as key and value.
Since the point-wise convolution kernel weights are fixed after training, the extraction in PWE is independent of the input image, \textit{i.e.,} content-agnostic. While, the attention mechanisms in SSE and CSE determine the mapping weights according to the similarity between channel slices, which are content-related. In addition, SSE and PWE contain the mapping between all the channel slice pairs of image feature and structure tokens, while the more efficient CSE only involves the one-way mapping of channel slices from image feature to structure tokens. 
Interestingly, we find that PWE shows more advantages compared with the other two counterparts under fewer extraction operations (more details please refer to Figure \ref{ablation_blocks}).
Furthermore, benefiting from stronger modeling capability, SSE begins to show its superiority under greater challenges. 


We evaluate our approach on three challenging semantic segmentation benchmarks under different backbones.
For example, equipped with ViT-L/16 \cite{VIT} as backbone, our StructToken achieves 54.18\% mIoU on ADE20K \cite{ADE20K}, 82.07\% mIoU on Cityscapes \cite{Cityscapes} and 49.07\% mIoU on COCO-Stuff-10K \cite{COCOSTUFF} respectively, which outperforms the state-of-the-art methods. 
Our main contributions include:
\begin{itemize}
\item We propose a new paradigm for semantic segmentation, termed structure-aware extraction paradigm, which follows a similar mechanism as humans and emphasizes the critical effect of structural information.
\item We present a network, named StructToken under our structure-aware extraction paradigm, and explore the different implementations of the extraction process.
\item Extensive experiments verify the effectiveness of our approach and show the prospect of human-like segmentation paradigm.
\end{itemize}

\section{Related Work}

\noindent\textbf{Static Per-pixel Classification Paradigm.}
Since Fully Convolutional Networks (FCN) \cite{FCN} were proposed, per-pixel classification has dominated semantic segmentation. 
It classifies each pixel to a specific category via a fixed classifier (such as a 1×1
convolution), which is unchangeable after the training process.
The methods under this paradigm mainly focused on how to learn better representation for each pixel via context modeling and fusion.
The early PSPNet \cite{PSPNet} used a pyramid pooling module to make multi-scale context fusion. 
The DeepLab family \cite{Deeplab,DeepLabv3} introduced the dilated convolution to expand the receptive fields. 
DANet \cite{DANet}, DSANet \cite{DSANet}, CCNet \cite{CCNet} and OCRNet \cite{OCRNet} used non-local modules to model more precise context information. 
\cite{tcsvt3} proposed a stage-aware feature alignment module to align and fusion of features between adjacent levels. \cite{tcsvt6} proposed the Gaussian dynamic convolution to adaptive fusion context information. \cite{tcsvt1} enhance the ability to locate object boundaries by cascaded CRFs. \cite{tcsvt7} extract the non-rigid geometry features by deformable convolution. \cite{tcsvt8} strengthens that connection to the same object through the object-level semantic integration module for more efficient integration of context information.
In addition, STLNet\cite{STLNet} starts to consider the structural information of the image itself, by introducing the texture enhance module and pyramid texture feature extraction Module to model the structural properties of textures in images. However, STLNet does it by modeling or statistically contextualizing the information itself, and it is still a per-pixel classification paradigm.
Recent work \cite{SETR,DPT,FTN,Segformer,FPT,feseformer} began to use transformer architecture to capture long-range context information.


\noindent\textbf{Dynamic Per-pixel Classification Paradigm.}
Compared with the static one,
this paradigm dynamically generates classifiers for each category based on the image content. Specifically, it establishes the connection between the image content and the classifier through attention and facilitates the classifier to be more suitable for the current sample image through the concatenation of multiple blocks.
Segmenter \cite{Segmenter} employed the transformer to jointly process the patches and class embeddings (tokens) during the decoding phase and let the class tokens perform matrix multiplication with the feature map to produce the final score map. MaskFormer \cite{Maskformer} unified instance segmentation and semantic segmentation architecture by performing matrix multiplication between class tokens and feature maps and using a binary matching mechanism. Mask2Former \cite{Mask2former} and K-Net \cite{knet} used learned semantic tokens, which are equivalent to the class tokens, to replace 1$\times$1 convolution kernels and used binary matching to unify semantic, instance, and panoptic segmentation tasks.



However, both of these two paradigms ignore the structural information of each category, which is critical in the human decision-making mechanism. 
Inspired by the segmentation discrimination mechanism of the human brain, we aim to explore a new paradigm to focus more on how to use the structural information as a cue for semantic segmentation task.

\section{Method}

In this section, we first present the overall framework of our StructToken. 
Then, we give the details about two basic components in each block of the decoder, the interaction module and the feed-forward network, respectively.
\begin{figure*}[tp]
\centering
\includegraphics[width=0.95\linewidth]{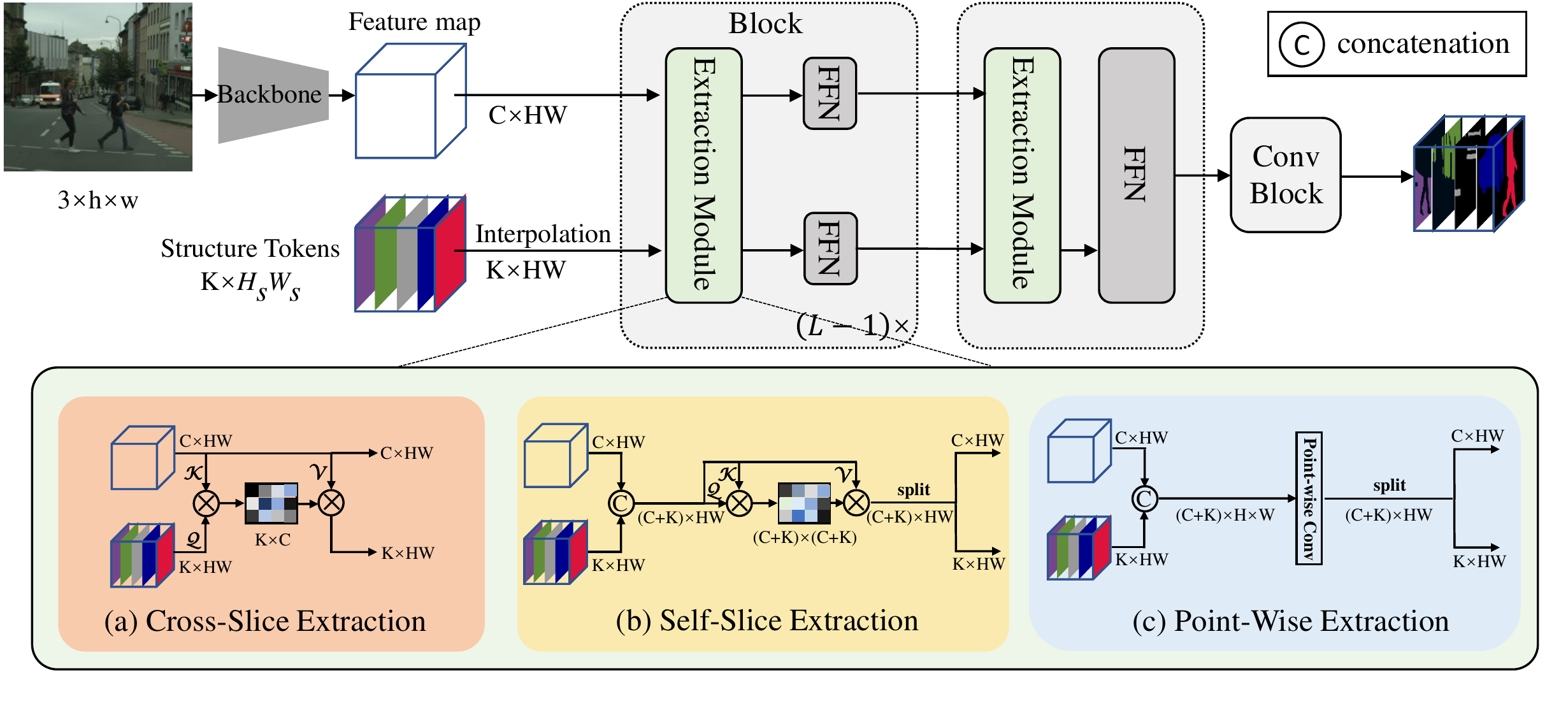} 
\caption{The overall framework of our StructToken. (a), (b) and (c) illustrate three different implementations of the extraction module, respectively. Here h and w represent the height and width of the original image, while H and W represent the height and width of the feature map output by backbone (e.g. using ViT\cite{VIT} as the backbone, the size of the output feature map is 1/16 of the original image). $H_S$ and $W_S$ represent the height and width of the structure t
okens. The $\mathcal{Q}$, $\mathcal{K}$ and $\mathcal{V}$ in CSE and SSE represent the query, key, and value output by the mapping functions $\bm{\Phi}$ and $\bm{\Psi}$, respectively. For more details, please see the method section.
}
\label{framework}
\end{figure*}
\subsection{Framework}

The overall framework of our StructToken is shown in Figure \ref{framework}.
Given an input image $\mathcal{I}$ $\in$ $\mathbb{R}^{3 \times h \times w}$, we first use a single-scale backbone (such as ViT \cite{VIT}) to generate the feature map $ \mathcal{F} \in \mathbb{R}^{C \times H \times W}$. $C$ is the channel number, and the $(h,w)$ and $(H,W)$ represent the spatial size (height and width) of the input image and feature map, respectively. 
Then, the feature map $\mathcal{F}$ and structure tokens $\mathcal{S} \in \mathbb{R}^{K \times H_s \times W_s}$ are sent to the decoder, where $K$ means the total number of categories within the dataset. 
The structure tokens are learnable during training and fixed during inference, each of which contains the implicit structural information of a specific category. Note that, when $(H,W) \neq (H_s,W_s)$, the structure tokens are interpolated to the same spatial size of the feature map $\mathcal{F}$.
The whole decoder contains $L$ consecutive blocks. Each block consists of an extraction module and two feed-forward networks (FFN).
The extraction module aims to extract the structural information from the feature map to structure tokens, and each of the resulting structure tokens can be regarded as a mask for each category.
The two FFN are used to refine the $\mathcal{F}$ and $\mathcal{S}$ via channel-wise projection respectively.
For the last block, only the FFN for structure tokens is required because the feature map is not used in the following process.

Finally, a simple ConvBlock \cite{identitymapping}, including two $3\times3$ convolutions and a skip connection, is applied to the output structure tokens of the last block to further refine the segmentation results.

\subsection{Extraction Module}

As the structure tokens are learned from the whole dataset, the structural information in them is abstract and implicit, which entails further specification and refinement according to the current input image.
Accordingly, the extraction module is designed to extract the structural information from the feature map to structure tokens.
For comprehensiveness, we explore using content-agnostic convolution and content-related attention to implement the extraction operation, resulting in three variants: point-wise extraction (PWE), self-slice extraction (SSE), and cross-slice extraction (CSE), respectively. 
Specifically, PWE and SSE apply $1\times1$ convolution and channel-wise self-attention to the concatenation of the image feature and structure tokens, respectively. CSE performs channel-wise cross-attention between structure tokens and image feature, where the query comes from the former and the latter is treated as key and value.
Since the extraction weights in PWE are independent of the input image, it is content-agnostic, while the attention-based SSE and CSE are content-related. In addition, CSE can be regarded as a simplified version of SSE with only one-way mapping of channel slices from image feature to structure tokens.
We provide the implementation details of these three variants as follows.

\subsubsection{Cross-Slice Extraction}
Considering that cross-attention is a well-known operation to aggregate information from one thing to another, which is highly compatible with the role of our interaction module. 
Thus, we utilize cross-attention to extract structural information from the feature map to structure tokens. 
Such process is named cross-slice extraction (CSE). 
The forward pass of CSE in the $i$-th block can be formulated as follows:
\begin{align} 
\label{eq:cse1}
& \mathcal{Q}_{i}=\bm{\Phi}_{q}(\mathcal{S}_{i}), \quad \mathcal{K}_{i}=\bm{\Phi}_{k}(\mathcal{F}_{i}), \quad
\mathcal{V}_{i}=\bm{\Phi}_{v}(\mathcal{F}_{i}), \\
\label{eq:cse2}
& \mathcal{S}_{i+1}=\text{Softmax}( \frac{ \mathcal{Q}_{i} \times \mathcal{K}_{i}^{T} }{\sqrt C} ) \times \mathcal{V}_{i}, 
\end{align}
where $\mathcal{F}_{i} \in \mathbb{R}^{C \times HW}$ is the input feature map, and $\mathcal{S}_{i} \in \mathbb{R}^{K \times HW}$ denotes the structure tokens, which can be viewed as $K$ tokens, each of which is a 2-dimension slice with height $H$ and width $W$. 
The query in cross-attention is generated by structure tokens, and the feature map is used to construct key and value.
As the original definition in \cite{nlpattn}, the projection layers $\bm{\Phi}_{\alpha \in \{q, k, v\}}$ here play a role of re-mapping each token in $\mathcal{S}_{i}$ and $\mathcal{F}_{i}$ via the same pattern.
However, simply performing a fully-connected layer along the $HW$ dimension on each token leads to the incompatibility for input images with arbitrary size as well as the multi-scale inference process.
In order to solve this problem, we replaced the three fully-connected projections with local-connected ones.
Specifically, the $\bm{\Phi}_{\alpha \in \{q, k, v\}}$ in Eq. \eqref{eq:cse1} is formulated as follows:
\begin{align} 
\label{eq:mapping}
\bm{\Phi}_{\alpha \in \{q, k, v\}} (x) = \bm \zeta_\alpha ( \bm\phi_\alpha( \bm \xi_\alpha(x))), 
\end{align}
where $\bm{\phi}_\alpha$ denotes a $3\times3$ depth-wise convolution which maps each token locally. $\bm{\zeta}_\alpha$ and $\bm{\xi}_\alpha$ are $1\times1$ point-wise convolution to make each token have a preview of its counterparts.






\subsubsection{Self-Slice Extraction}

In this variant, we use self-attention to interact structure tokens and the feature map with each other. 
Specifically, in the forward process of the $i$-th block, it first concatenates the structure tokens $\mathcal{S}_{i} \in \mathbb{R}^{K \times HW}$ and the feature map $\mathcal{F}_{i} \in \mathbb{R}^{C \times HW}$ along the channel dimension,
\begin{align} 
\label{eq:sse1}
& \mathcal{G}_{i}=\text{Concat}(\mathcal{S}_{i}, \mathcal{F}_{i}) \in \mathbb{R}^{(C+K) \times HW}, 
\end{align} 
Then, the self-attention is performed on $\mathcal{G}_{i}$ to exchange information between structure tokens and feature map,
\begin{align} 
\label{eq:sse2}
& \mathcal{Q}_{i}=\bm{\Psi}_{q}(\mathcal{G}_{i}), \quad \mathcal{K}_{i}=\bm{\Psi}_{k}(\mathcal{G}_{i}), \quad
\mathcal{V}_{i} = \bm{\Psi}_{v}(\mathcal{G}_{i}), \\
& \widehat{\mathcal{G}}_{i}=\text{Softmax}( \frac{ \mathcal{Q}_{i} \times \mathcal{K}_{i}^{T} }{\sqrt C} ) \times \mathcal{V}_{i}, 
\end{align} 
where, the projection layer $\bm{\Psi}_{\alpha \in \{q,k,v\}}$ share the same implementation of the $\bm{\Phi}_{\alpha \in \{q,k,v\}}$ in CSE.
$\mathcal{Q}_{i}$, $\mathcal{K}_{i}$ and $\mathcal{V}_{i}$ have the same shape with $(C+K) \times HW$.
Finally, the structure tokens $\mathcal{S}_{i+1} \in \mathbb{R}^{K \times HW}$ and feature map $\mathcal{F}_{i+1} \in \mathbb{R}^{C \times HW}$ are divided from the updated $\widehat{\mathcal{G}}_{i}$ by directly split along the channel dimension,
\begin{align} 
\label{eq:sse3}
& \mathcal{S}_{i+1}, \mathcal{F}_{i+1} = \text{Split} (\widehat{\mathcal{G}}_{i}). 
\end{align} 
It can be found that the interaction in CSE is uni-directional ($\mathcal{S} \rightarrow \mathcal{F}$) with only structure tokens being updated, while our SSE achieves a more comprehensive bi-directional interaction ($\mathcal{S} \leftrightarrow \mathcal{F}$) in which both structure tokens and feature map are updated.
Thus, SSE can be regarded as an extension of the CSE.


\subsubsection{Point-Wise Extraction}
As stated above, the attention map (with shape $\mathbb{R}^{(C+K) \times (C+K)}$) in SSE represents the aggregation weights for every slice of the concatenated feature, which is further used to filter out the unuseful information.
Different from using dot-product to generate the aggregation weights, our point-wise extraction (PWE) is designed to directly learn the weights via a simple point-wise convolution layer.
To be specific, in the forward process of the $i$-th decoder block, we also first concatenate the structure tokens $\mathcal{S}_{i} \in \mathbb{R}^{K \times HW}$ and the feature map $\mathcal{F}_{i} \in \mathbb{R}^{C \times HW}$ according to Eq. \eqref{eq:sse1}, resulting in $\mathcal{G}_{i} \in \mathbb{R}^{(C+K) \times HW}$.
Then, the interaction is performed via the point-wise convolution $\bm \Omega$, whose parameters are deemed the aggregation weights,
\begin{align} 
\label{eq:pwe}
& \tilde{\mathcal{G}}_{i}=\bm \Upsilon(\mathcal{G}_{i}) \in \mathbb{R}^{(C+K) \times HW}, \\
& \widehat{\mathcal{G}}_{i}=\bm \Omega (\tilde{\mathcal{G}}_{i}) \in \mathbb{R}^{(C+K) \times HW},
\end{align} 
where the projection layer $\bm \Upsilon$ is implemented same as the $\bm{\Psi}$ and $\bm \Phi$ in Eq. \eqref{eq:cse1} and Eq. \eqref{eq:sse2}. 
The $\bm \Omega$ denotes the point-wise convolution.


\subsection{Feed-Forward Networks (FFN)}

The traditional feed-forward networks (FFN) \cite{nlpattn} is comprised of two consecutive fully connected layers to expand and shrink the channel dimension respectively.
Considering that the FFN in our framework plays a role of refinement, we added a lightweight 3$\times$3 group convolution \cite{alexnet} between the original two fully connected layers to involve more local context (ablated in Table \ref{addmodule}). 


\section{Experiments}

We first introduce the datasets and implementation details.
Then, we compare our method with the recent state-of-the-arts on three challenging semantic segmentation benchmarks.
Finally, comprehensive ablation studies and visual analysis are conducted to evaluate the effectiveness of our approach.

\subsection{Datasets}

\noindent \textbf{ADE20K} \cite{ADE20K} is a challenging scene parsing dataset, which is split into 20210 and 2000 images for training and validation, respectively. It has 150 fine-grained object categories and diverse scenes with 1,038 image-level labels.

\noindent \textbf{Cityscapes} \cite{Cityscapes} carefully annotates 19 object categories of urban driveway landscape images. It contains 5K finely annotated images and is divided into 2975 and 500 images for training and validation, respectively. It is a high-quality dataset.

\noindent \textbf{COCO-Stuff-10K} \cite{COCOSTUFF} is a significant scene parsing benchmark with 9000 training images and 1000 testing images. It has 171 categories.

\begin{table}[t]
\setlength\tabcolsep{2pt}
\centering
\caption{Comparison with the state-of-the-art methods on the ADE20K dataset. ``SS'' and ``MS'' indicate single-scale and multi-scale inference, respectively. $^{\dag}$ means the ViT models trained from scratch on ImageNet-21k and fine-tuned on ImageNet-1k \cite{vitpretrain}. $^{*}$ represents our implementation under the same settings as the official repo.}

  \begin{tabular}{l|c|c|c|c|c|c}
    \hline
    \hline
    \multirow{2}{*}{Method} & \multirow{2}{*}{Venue} & \multirow{2}{*}{Backbone} & \multirow{2}{*}{GFLOPs} & \multirow{2}{*}{Params} & {mIoU} & {mIoU}\\
    & & & & &(SS)&(MS)\\
    \hline
    FCN \cite{FCN}            &CVPR15 & ResNet-101  &276 &69M &39.91 & 41.40 \\
    EncNet \cite{EncNet}      &CVPR18 & ResNet-101  &219 &55M &-     & 44.65 \\
    OCRNet \cite{OCRNet}      &ECCV20 & HRNet-W48   &165 &71M &43.25 & 44.88 \\
    CCNet \cite{CCNet}        &ICCV19 & ResNet-101  &278 &69M &43.71 & 45.04 \\
    ANN \cite{ANN}            &ICCV19 & ResNet-101  &263 &65M &-     & 45.24 \\
    PSPNet \cite{PSPNet}      &CVPR17 & ResNet-101  &256 &68M &44.39 & 45.35 \\
    FPT \cite{FPT}            &ECCV20 & ResNet-101  &-   &-   &-     & 45.90\\
    DeepLabV3+ \cite{v3+}     &ECCV18 & ResNet-101  &255 &63M &45.47 & 46.35 \\
    STLNet\cite{STLNet}       &CVPR21 & ResNet-101  &-   &-   &-     &46.48 \\
    DMNet \cite{DMNet}        &ICCV19 & ResNet-101  &274 &72M &45.42 & 46.76\\
    ISNet \cite{ISNet}        &ICCV21 & ResNeSt-101 &-   &-   &-     & 47.55\\
    \hline
    DPT \cite{DPT}            &ICCV21 & ViT-Hybrid         &-   &-    &-     & 49.02\\
    DPT$^{*}$                 &ICCV21 & ViT-L/16$^{\dag}$       &328 &338M &49.16 & 49.52\\
    UperNet$^{*}$             &ECCV18 & ViT-L/16$^{\dag}$       &710 &354M &48.64 & 50.00\\
    SETR \cite{SETR}          &CVPR21 & ViT-L/16          &214 &310M &48.64 & 50.28\\
    MCIBI \cite{MCIBI}        &ICCV21 & ViT-L/16                 &-   &-    &-     & 50.80\\
    SegFormer \cite{Segformer}&NeurIPS21 & MiT-B5                 &183 &85M  &-     & 51.80\\
    SETR-MLA$^{*}$            &CVPR21 & ViT-L/16$^{\dag}$      &214 &310M &50.45 & 52.06\\
    UperNet \cite{swin}       &ECCV18 & Swin-L                 &647 &234M &52.10 & 53.50\\
    Segmenter \cite{Segmenter}&ICCV21 & ViT-L/16$^{\dag}$      &380 &342M &51.80 & 53.60\\
    \rowcolor{gray}
    StructToken-SSE  &- & ViT-L/16$^{\dag}$      &486 &395M & 52.82 & 54.00\\
    \rowcolor{gray}
    StructToken-CSE  &- & ViT-L/16$^{\dag}$      &398 &350M &52.84 & \textbf{54.18}\\
    \rowcolor{gray}
    StructToken-PWE  &- & ViT-L/16$^{\dag}$      &442 &379M &\textbf{52.95} & 54.03\\
    \hline
    \hline
    \end{tabular}
\label{sotaade}
\end{table}

\begin{table}[!h]
\setlength\tabcolsep{1.8pt}
\centering
\caption{Comparison with the state-of-the-art methods on the Cityscapes validation set.}
  \begin{tabular}{l|c|c|c|c|c|c}
    \hline
    \hline
    \multirow{2}{*}{Method}& \multirow{2}{*}{Venue} & \multirow{2}{*}{Backbone} & \multirow{2}{*}{GFLOPs} & \multirow{2}{*}{Params} & {mIoU} & {mIoU}\\
    & & & & &(SS)&(MS)\\
    \hline
    FCN \cite{FCN}            &CVPR15   & ResNet-101 &633 &69M & 75.52 & 76.61 \\
    EncNet \cite{EncNet}      &CVPR18   & ResNet-101 &502 &55M & 76.10 & 76.97 \\
    PSPNet \cite{PSPNet}      &CVPR17   & ResNet-101 &585 &68M & 78.87 & 80.04 \\
    GCNet \cite{GCNet}        &ICCVW19   & ResNet-101 &632 &69M & 79.18 & 80.71 \\
    DNLNet \cite{DNL}         &ECCV20   & ResNet-101 &637 &69M & 79.41 & 80.68 \\
    CCNet \cite{CCNet}        &ICCV19   & ResNet-101 &639 &69M & 79.45 & 80.66 \\
    \hline
    Segmenter \cite{Segmenter} &ICCV21  & DeiT-B     &-   &-    &79.00  & 80.60\\
    Segmenter \cite{Segmenter} &ICCV21  & ViT-L/16      &553 &340M &79.10  & 81.30\\
    \rowcolor{gray}
    StructToken-CSE   &- & ViT-L/16      &567 &349M & 79.64 & 81.98 \\
    \rowcolor{gray}
    StructToken-SSE   &- & ViT-L/16      &651 &377M & 80.01 & 82.02 \\
    \rowcolor{gray}
    StructToken-PWE   &- & ViT-L/16      &600 &364M &\textbf{80.05} & \textbf{82.07}\\
    \hline
    \hline
    \end{tabular}
    \label{sotacitys}
\end{table}

\subsection{Implementation Details}
All the experiments are conducted on 8 NVIDIA Tesla V100 GPUs (32 GB memory per-card) with PyTorch implement and mmsegmentation\cite{mmsegmentation} codebase.
We use ViT \cite{VIT} as the backbone.
During training, we follow the common setting using data augmentation such as random horizontal flipping, random resize, random cropping ($512 \times 512$ for ADE20K and COCO-Stuff-10K, $768\times 768$ for Cityscapes and 640$\times$640 with ViT-L/16 for ADE20K), etc. As for optimization, we adopt a polynomial learning rate decay schedule; following prior works \cite{swin}, we employ AdamW to optimize our model with 0.9 momenta and 0.01 weight decay; we set the initial learning rate as 2e-5. The batch size is set to 16 for all datasets. The total iterations are 160k, 80k, and 80k for ADE20K, Cityscapes and COCO-Stuff-10K, respectively. 
For inference, we follow previous work \cite{swin,SETR} to average the multi-scale (0.5, 0.75, 1.0, 1.25, 1.5, 1.75) predictions of our model. Interpolation operations are used for multi-scale inference. The slide-window test is applied here. The performance is measured by the widely-used mean intersection of union (mIoU) in all experiments. 
Considering the effectiveness and efficiency, we adopt the ViT-T/16 \cite{VIT} as the backbone in the ablation study on ADE20K.

\subsection{Comparisons with the State-of-the-art Methods}

\noindent\subsubsection{Results on ADE20K} 
Table \ref{sotaade} reports the comparison with the state-of-the-art methods on the ADE20K validation set. 
From these results, it can be seen that our StructToken is +1.02$\%$, +1.15$\%$ and +1.04$\%$ mIoU (52.82, 52.95 and 52.84 vs. 51.80) higher than Segmenter \cite{Segmenter} with the same input size (640$\times$640), respectively.
When multi-scale testing is adopted, our StructToken is +0.4$\%$, +0.43$\%$ and +0.58$\%$ mIoU (54.00, 54.03 and 54.18 vs. 53.60) higher than Segmenter, respectively.
For ViT-T/16, as shown in Table \ref{diffbackbone}, our best results is +0.86$\%$ mIoU (42.99 vs. 42.13) higher than DPT \cite{DPT} with the same input size (512$\times$512). For ViT-S/16, our best result is +1.44$\%$ mIoU (48.89 vs. 47.45) higher than DPT. For ViT-B/16, our best result is +1.82$\%$ mIoU (51.82 vs. 50.00) higher than Segmenter. Furthermore, the larger the model is, the better StructToken performs.

\noindent\subsubsection{Results on Cityscapes} 
Table \ref{sotacitys} demonstrates the comparison results on the validation set of Cityscapes. The previous state-of-the-art method Segmenter with ViT-L/16 achieves 79.10$\%$ mIoU. Our StructToken is +0.54$\%$, +0.91$\%$ and +0.95$\%$ mIoU (79.64, 80.01 and 80.05 vs. 79.10) higher than it, respectively. As to multi-scale inference, our method is +0.68$\%$, +0.72$\%$ and +0.77$\%$ mIoU (81.98, 82.02, 82.07 vs. 81.30) higher than Segmenter, respectively.

\noindent\subsubsection{Results on COCO-Stuff-10K}
Table \ref{sotacoco} compares the segmentation results on the COCO-Stuff-10K testing set. It can be seen that our StructToken-SSE can
achieve $49.07\%$ mIoU, and our method is +4.18$\%$ mIoU higher than MCIBI \cite{MCIBI} (49.07 vs. 44.89).

\begin{table*}[t]
\centering
\setlength\tabcolsep{19pt}
\caption{Ablation study of each component in our StructToken on ADE20K. ``$\diamondsuit$'' means the basic architecture of FFN, \textit{i.e.,} two consecutive linear layers, and ``$\diamondsuit \spadesuit$'' denotes the above basic FFN with a $3 \times 3$ group convolution between two linear layers to enhance the locality.
     All the experiments are equipped with ViT-T/16 as the backbone.}
    \begin{tabular}{ccc|c|c|c|c}
     \hline
     \hline
     \multirow{2}{*}{Interaction Module}   & \multirow{2}{*}{FFN}  & \multirow{2}{*}{ConvBlock} & \multirow{2}{*}{GFLOPs} & \multirow{2}{*}{Params}  & {mIoU} & {mIoU}\\
     & & & & & (SS) &(MS)\\
     \hline
     CSE  & $\diamondsuit$    &              & {6.74}      & {8.5M}         & {37.71}    & {38.90}   \\
     CSE  & $\diamondsuit$ $\spadesuit$  &              & {6.74}      & {8.5M}         & {38.17}    & {38.85}   \\
     CSE  & $\diamondsuit$           & $\checkmark$         & {7.16}      & {8.9M}         & {38.01}    & {39.29}   \\
     CSE  & $\diamondsuit$ $\spadesuit$      & $\checkmark$  & {7.16}    & {8.9M}  & \textbf{39.12}  & \textbf{40.23}  \\
     \hline
     \hline
     \end{tabular}
     \label{addmodule}
\end{table*}

\begin{table}[t]
\setlength\tabcolsep{11.9pt}
\centering
\caption{Comparison with the state-of-the-art methods on the COCO-Stuff-10K dataset.}
  \begin{tabular}{l|c|c|c}
    \hline
    \hline
    \multirow{2}{*}{Method} &\multirow{2}{*}{Venue} & \multirow{2}{*}{Backbone}    & mIoU \\
    &  & & (MS) \\
    \hline
    PSPNet \cite{PSPNet}         &CVPR17 & ResNet-101 & 38.86 \\
    SVCNet \cite{SVCNet}         &CVPR19 & ResNet-101 & 39.60 \\
    DANet \cite{DANet}           &CVPR19 & ResNet-101 & 39.70 \\
    EMANet \cite{EMANet}         &ICCV19 & ResNet-101 & 39.90 \\
    SpyGR \cite{SpyGR}           &CVPR20 & ResNet-101 & 39.90 \\
    ACNet \cite{ACNet}           &ICCV19 & ResNet-101 & 40.10 \\
    OCRNet \cite{OCRNet}         &ECCV20 & HRNet-W48 & 40.50 \\
    GINet \cite{GINet}           &ECCV20 & ResNet-101 & 40.60 \\
    RecoNet \cite{RecoNet}       &ECCV20 & ResNet-101 & 41.50\\
    ISNet \cite{ISNet}           &ICCV21 & ResNeSt-101& 42.08\\
    \hline
    MCIBI \cite{MCIBI}          &ICCV21 & ViT-L/16      & 44.89\\
    \rowcolor{gray}
    StructToken-PWE    &- & ViT-L/16      & 48.24\\
    \rowcolor{gray}
    StructToken-CSE    &- & ViT-L/16      & 48.71 \\
    \rowcolor{gray}
    StructToken-SSE    &- & ViT-L/16      & \textbf{49.07}\\
    \hline
    \hline
    \end{tabular}
    \label{sotacoco}
\end{table}

\subsection{Ablation Study}
In this section, all the models in the following experiments adopt ViT-T/16 \cite{VIT} as the backbone and are trained on ADE20K training set for 160K iterations.
Our baseline model is the CSE module and FFN module without grouped convolution. Note that we does not perform ablation experiments using a fully connected layer to map query, key, and value matrices because it does not support the multi-scale inference.

\subsubsection{Effect of Each Component}
As shown in Table \ref{addmodule}, we experiment with adding a 3$\times$3 group convolution layer \cite{alexnet} to the FFN module and a ConvBlock to the model. In addition, the FLOPs of FFN with a group convolution layer are only 0.002G, which is ignored in Table \ref{addmodule}. It is a lightweight convolution layer, and the performance of the model reaches 39.12$\%$ mIoU after the FFN module and ConvBlock module are added, which is +1.41$\%$ mIoU (39.12 vs. 37.71) higher than the base model, and +1.33 $\%$ mIoU (40.23 vs. 38.90) for multi-scale inference.

\begin{table}[t]
\newcommand{\tabincell}[2]{\begin{tabular}{@{}#1@{}}#2\end{tabular}}
\setlength\tabcolsep{5.5pt}
\centering
\caption{Compare the performance of ViT variants on the ADE20K dataset.}
  \begin{tabular}{l|c|cccc}
    \hline
    \hline
    \multirow{2}{*}{Method}  & \multirow{2}{*}{Backbone}    & \multirow{2}{*}{GFLOPs}      & \multirow{2}{*}{Params}  & mIoU & mIoU \\
    & & & & (SS) & (MS) \\
    \hline
    Segmenter                       &              &6   &7M    &38.10     & 38.80\\
    UperNet                         &              &35  &11M   &38.93     & 39.19\\
    SETR-MLA                        &              &10  &11    &39.88     & 41.09\\
    DPT                             & ViT-T/16     &104 &17M   &40.82     & 42.13\\
    \rowcolor{gray}
    StructToken-CSE        &              &7   &9M    &39.12     & 40.23\\
    \rowcolor{gray}
    StructToken-SSE        &              &13  &14M   & 40.81  & 42.24 \\
    \rowcolor{gray}
    StructToken-PWE        &              &10  &12M   &\textbf{41.87}  & \textbf{42.99}\\
    \hline
    UperNet                         &              &140 &42M   &45.53 & 46.14\\
    SETR-MLA                        &              &21  &27M   &44.85 & 46.30\\
    Segmenter                       &              &22  &27M   &45.00 & 46.90\\
    DPT                             & ViT-S/16     &118 &36M   &46.37 & 47.45\\
    \rowcolor{gray}
    StructToken-CSE        &              &23  &30M   &45.86 & 47.44\\
    \rowcolor{gray}
    StructToken-SSE        &              &37  &41M   &47.11 & \textbf{49.07}\\
    \rowcolor{gray}
    StructToken-PWE        &              &31  &38M   &\textbf{47.36} & 48.89\\
    \hline
    UperNet                         &              &292 &128M  &46.58 & 47.47\\
    DPT                             &              &171 &110M  &47.20 & 47.86\\
    SETR-MLA                        &              &65  &92M   &48.21 & 49.32\\
    Segmenter                       & ViT-B/16     &81  &107M  &49.00 & 50.00\\
    \rowcolor{gray}
    StructToken-CSE        &              &86  &113M  &49.51 & 50.87\\
    \rowcolor{gray}
    StructToken-SSE        &              &123 &142M  &50.72 & \textbf{51.85} \\
    \rowcolor{gray}
    StructToken-PWE        &              &105 &132M  &\textbf{50.92} & 51.82\\
    \hline
    \hline
    \end{tabular}
    \label{diffbackbone}
\end{table}


\begin{figure}[h]
\centering
\includegraphics[width=1\linewidth]{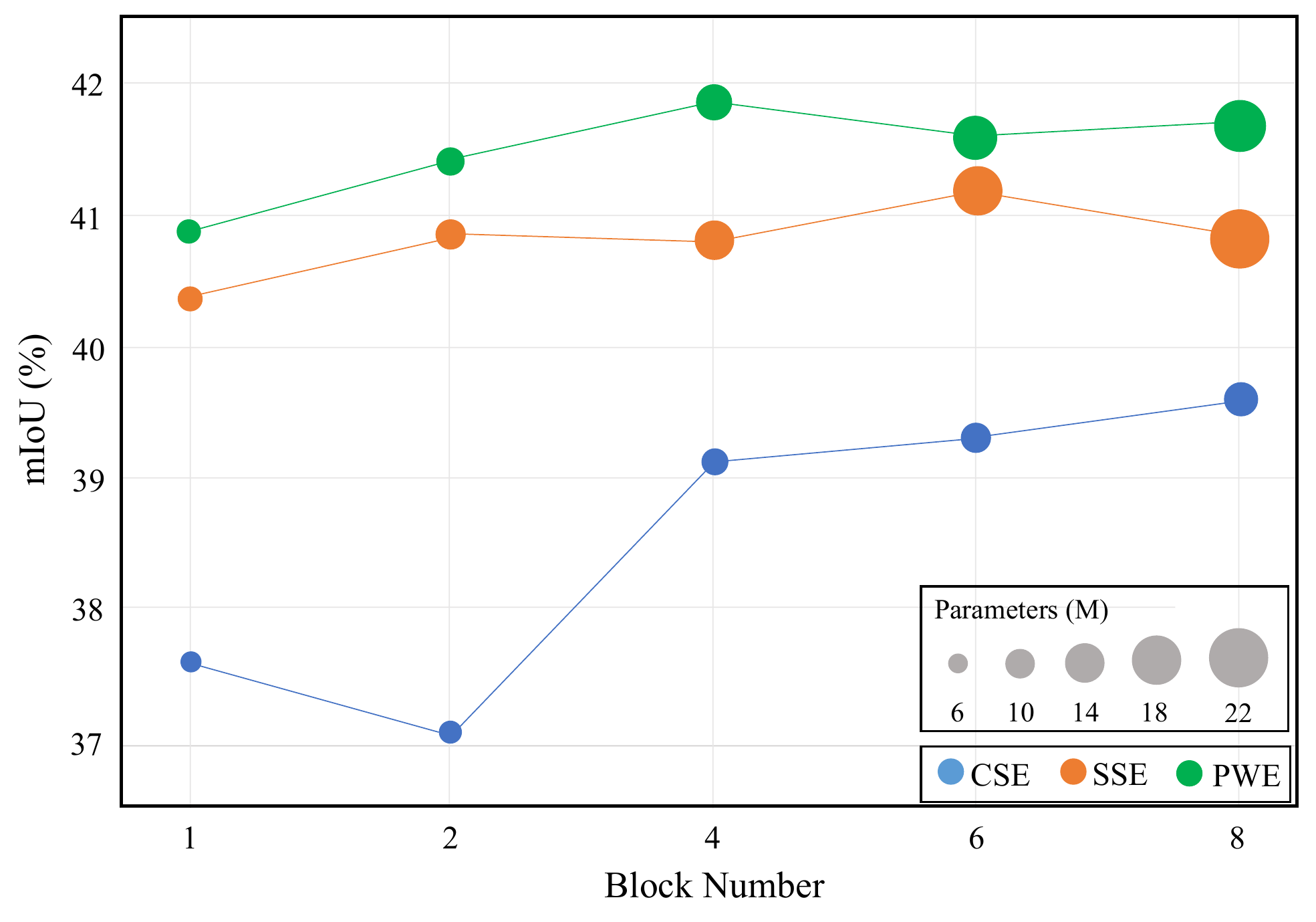} 
\caption{Comparison under different block numbers on ADE20K. Here, ViT-T/16 is used as the backbone.}
\label{ablation_blocks}
\end{figure}

\subsubsection{Number of Blocks}


Figure \ref{ablation_blocks} shows the comparison among StructToken-CSE, StructToken-SSE and StructToken-PWE under different block numbers.
It can be seen that the performance of all the variants presents an upward trend with the increase of block number. For the trade-off between performance and computation complexity and the number of parameters, we choose to use 4 blocks by default for all the variants, which also means that the performance of our StructToken in Table \ref{sotaade},\ref{sotacitys} and \ref{sotacoco} are lower than its upper bound.
Interestingly, SSE and CSE with more flexible content-related attention operation perform worse than the content-agnostic PWE, and performance gap between them narrows with the increase of block number.
This may be attributed to the more inflexible convolution operation easier to learn, while the attention operation need more blocks to show its strengths.


\subsubsection{Comparison of CSE, SSE and PWE}
\begin{figure*}[t]
\centering
\includegraphics[width=1\linewidth]{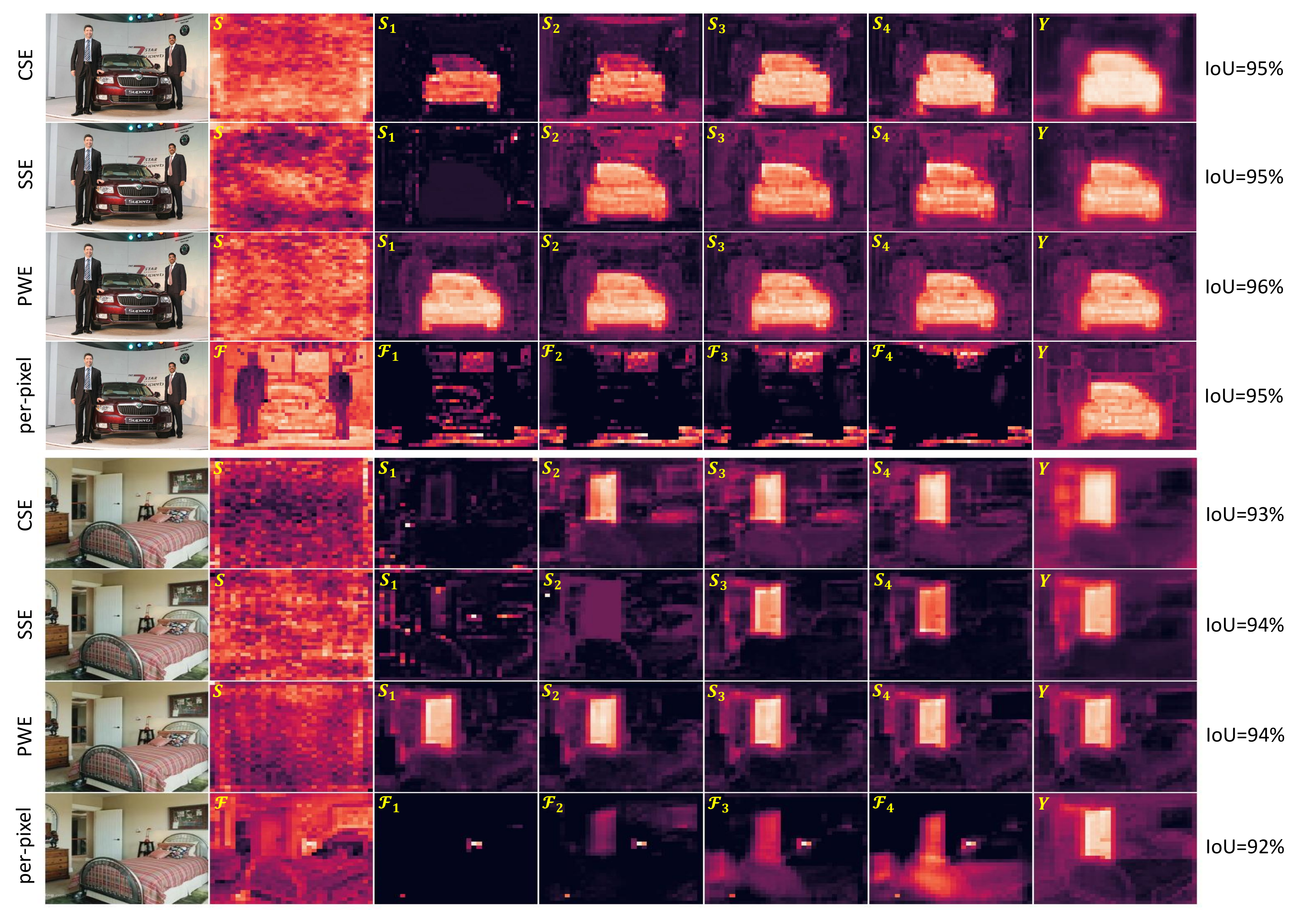} 
\caption{
Visualization of our three variants following structure-aware extraction paradigm (row 1$\sim$3) and their counterpart following per-pixel classification paradigm (row 4).
We choose two examples (1$^{st}$ column) including ``car'' and ``door'' category respectively.
For the 1$^{st}\sim$ 3$^{rd}$ row of each example, the $S$ in the second column denotes the structure tokens learned from the dataset, which contains the implicit structural information of each category. $S_i$ in 3$^{rd}\sim$6$^{th}$ columns represent the output structure tokens of the $i^{th}$ block.
The $Y$ in the last column indicates the output score map in the final layer.
For the 4$^{th}$ row of each example, the $\mathcal{F}$ in the second column denotes the backbone output feature, and $\mathcal{F}_i$ in the 3$^{rd}\sim$6$^{th}$ columns represent the output feature of $i^{th}$ residual block.
``IoU'' means the intersect over union score of the specific class (``car'' or ``door'') in the image.
}
\label{visual}
\end{figure*}

We compare these three variants from the following three aspects:
(a) \textit{The complexity of scenarios}. 
As can be seen from Table \ref{sotaade}, \ref{sotacitys} and \ref{sotacoco}, StructToken-PWE tends to perform better under small dataset and simple scenario (e.g., cityscapes with 19 categories).
For the moderately complex scenarios (such as ADE20k with 150 classes), SSE, CSE and PWE have similar performanc, in addition CSE saves 18$\%$ GFLOPs compared to SSE.
However, as the scenario becomes more complicated (e.g., COCO-Stuff-10K with 171 categories), 
content-related attention (i.e., SSE and CSE) begins to show its strengths, having benefited from dynamic modeling. The SSE with greater complexity performs better in this case. Compared to SSE and CSE, while PWE performs poorly on larger datasets, it performs well on smaller ones.
(b) \textit{Strength of backbone}. 
As shown in Table \ref{diffbackbone}, when using ViT-T/16\cite{VIT} as the backbone, StructToken-PWE surpasses CSE and SSE counterparts by a large margin, with +2.75$\%$ and +1.06$\%$ mIoU respectively.
As the backbone gets stronger, such performance gap gradually narrows (StructToken-PWE is only +0.2$\%$ mIoU higher than StructToken-SSE), and the content-relevant SSE gradually shows its advantages. 
In addition, content-relevant attention is more dependent on the features extracted by the backbone, and the richer the features, the better the performance.
(c) \textit{Number of decoder blocks}. 
The quantitative results in Figure \ref{ablation_blocks} and qualitative visualization in Figure \ref{visual} show that PWE has a stronger advantage with few decoder blocks. The performance gap is reduced with the increase of block number.

\subsection{Visual Analysis}



To better understand the mechanism of our paradigm, we visualize the evolution of structure tokens with progressive extraction operations to show how it works.
Figure \ref{visual} shows two examples sampled from ADE20K dataset, containing ``car'' and ``door'' respectively. 
We compare the three extraction methods in the first three rows of each example, where the 2$^{nd}$ column denotes the learned structure token slice corresponding to a specific category (``car'' or ``door'') and the 3$^{rd}\sim$6$^{th}$ columns represent its updated results in 1$^{st}\sim$4$^{th}$ blocks. 
From the 2$^{nd}$ column, we can find that the structure token learned from dataset is relatively abstract. It does not present an obvious object pattern, which is also understandable because of the diversity of objects in each category. 
The 3$^{rd}\sim$6$^{th}$ columns show that the structural information in the structure token is more and more obvious with the gradual extraction operations.
In addition, StructToken-PWE presents a rough object outline after the first block (3$^{rd}$ column), while such phenomenon is much more ambiguous in the other two counterparts, which means that PWE can extract information from the image feature much faster. In contrast, the extraction process of content-related SSE seems the slowest.

We further visually compare our paradigms and per-pixel classification paradigms to better understand the differences in how they work.
For better comparison, we instantiate a model following the per-pixel classification paradigm as a counterpart, which is more aligned with our paradigm. Specifically, we first apply a $1\times1$ convolution on the backbone output feature to project the channel number to the category number, then use four residual blocks \cite{ResNet} to transform the feature map, followed by a $1\times1$ convolution to generate the final segmentation result. 
So the feature map output of each residual block has a similar meaning to structure tokens, \textit{i.e.}, each slice contains the structure information of a specific category.
But their difference is that the structural information in the per-pixel classification paradigm only comes from the current input image, while the structural information in structure tokens is the prior knowledge learned from the dataset. 
In Figure \ref{visual}, the 4$^{th}$ row of each example shows the visualization of the feature slice corresponding to the specific category from each residual block output.
We can find that even though the per-pixel classification paradigm is similar with our paradigm in the final mIoU and output feature of the segmentation head, the feature map after each block presents a completely different pattern compared with the structure tokens. 
From 1$^{st}\sim$3$^{rd}$ rows, we can see the clear structure of the ``car'' and ``door'' categories in the structure tokens. In contrast, in the 4$^{th}$ row, we can only see the blurry structure or even no structure of the semantic class until the $1\times1$ convolution transform the feature maps to per-pixel classification score map. 
Such more explicit structure information provides strong evidence of the strength of our paradigm in retaining structural information.


\section{Conclusion}
In this paper, we propose a new paradigm different from the per-pixel classification, termed structure-aware extraction. The classical per-pixel classification methods only focus on learning better pixel representations or classification kernels while ignoring the structural information of objects, which is critical to human decision-making mechanism. In contrast, structure-aware extraction has a good ability to extract structural features. Specifically, it generates the segmentation results via the interactions between a set of learned structure tokens and the image feature, which aims to progressively extract the structural information of each category from the feature. We hope this work can bring some fundamental enlightenment to semantic segmentation and other tasks.

\section{Acknowledgements}
This research is supported by the National Natural Science Foundation of China [grant number U2003208], the Xinjiang Autonomous Region key research and development project [grant number 2021B01002] and The Xinjiang Autonomous Region major scientific and technological projects
[grant number 2020A03004-4].

{\small
\bibliographystyle{templete/ieee_fullname}
\bibliography{egbib}
}

\vfill

\end{CJK}
\end{document}